\pdfoutput=1
\documentclass[11pt]{article}
\usepackage[]{acl}
\usepackage{times}
\usepackage{latexsym}
\usepackage{todonotes}
\usepackage{amssymb,amsthm,mathtools,stackrel}
\usepackage{subfigure}
\usepackage{enumitem}

\usepackage[T1]{fontenc}
\usepackage[utf8]{inputenc}

\usepackage{microtype}
\usepackage{graphicx}
\usepackage{xr}

\newcommand{\nidhi}[1]{{\color{blue}{[N: #1]}}}

\newcommand{\cut}[1]{}
\newcommand{\abr}[1]{\textsc{#1}}
\newcommand{\gtnn}[0]{\abr{gtnn}}
\newcommand{\wikipedia}[0]{\abr{wikipedia}}
\newcommand{\pgr}[0]{\abr{pgr}}
\newcommand{\gdpr}[0]{\abr{gdpr}}

\def\bh{{\mathbf h}} 
\def\bx{{\mathbf x}} 
\def\bz{{\mathbf z}} 
\def\ba{{\mathbf a}} 
\def\bb{{\mathbf b}} 
\def\bW{{\mathbf W}}

\title{Generic and Trend-aware Curriculum Learning for Relation Extraction in Graph Neural Networks}

\author{Nidhi Vakil \\
  Department of Computer Science \\
  University of Massachusetts Lowell \\
  \texttt{nvakil@cs.uml.edu} \\ \And
  Hadi Amiri \\
  Department of Computer Science \\
  University of Massachusetts Lowell \\
  \texttt{hadi@cs.uml.edu} \\}

\begin{document}
\maketitle

\begin{abstract}
% We present a generic and trend-aware curriculum learning approach that effectively integrates textual and structural information in text graphs for relation extraction between entities, which we consider as node pairs in graphs. 
We present a generic and trend-aware curriculum learning approach for graph neural networks. It extends existing approaches by incorporating sample-level loss trends to better discriminate easier from harder samples and schedule them for training. The model effectively integrates textual and structural information for relation extraction in text graphs. Experimental results show that the model provides robust estimations of sample difficulty and shows sizable improvement over the state-of-the-art approaches across several datasets. 
%compared to the state-of-the-art curriculum learning approaches 

%resulting in an average of 9.3 points improvement in F1 score over the best-performing baseline model. 

%2.4 to 6.5 points

% In this research, we investigate the task of detecting relations/links between pair of entities given their detailed textual description and the network. We developed a Graph Text Neural Network (GTNN) model that effectively integrates textual and structural information to predict relations between the pair of entities in a supervised settings.
% Taking GTNN model, we further improved our model by curriculum learning on edge prediction. We extend the existing dynamic curriculum framework, SuperLoss (SL) which works on the top of the model loss. During the training, SL calculates the confidence of the instances based on the threshold at \textit{batch-level} to classify an instance into easy/hard category.We extend this framework at \textit{instance-level} to incorporate the information from the history of losses for each instance to captures the rising/falling loss trend in SL framework. We call our novel framework Trend-SL (Trend-SuperLoss). Our model, \gtnn{} with Trend-SL shows sizable improvement over the state-of-the-art Transformer-based approach on a publicly available datasets resulting in 17.57 and 6.47 points improvement in F1 score over the best baseline model respectively. 
\end{abstract}
\section{Introduction}

\begin{figure}
    \centering
    \includegraphics[scale=0.66]{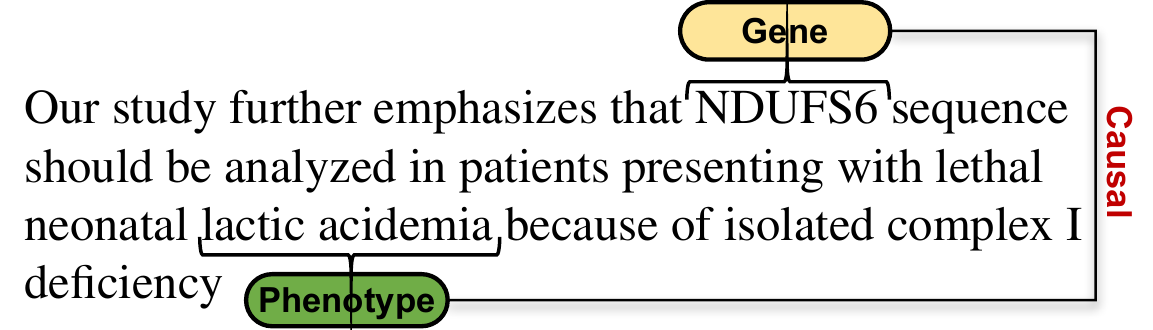}
    \caption{An example showing the report of a causal relation between a gene and a phenotype (symptom) from the PGR dataset~\citep{sousa2019silver}.}
    \label{fig:pgr_example_intro}
\end{figure}

Relation extraction is the task of detecting (often pre-defined) relations between entity pairs. It has been investigated in both natural language processing~\cite{mintz-etal-2009-distant,lin-etal-2016-neural,peng2017cross,zhang-etal-2018-graph} and network science~\cite{zhang2018link,alex2017protein}. Relation extraction is a challenging task, especially when data is scarce. %For example, in medical domain where a clinician needs to decide existence of a relation for a given entity pair like gene and disease, if the gene can cause the disease. 
Nonetheless, the ability to automatically link entity pairs is a crucial task as it can reveal relations that have not been previously identified, e.g., informing clinicians about a causal relation between a gene and a phenotype or disease. Figure~\ref{fig:pgr_example_intro} shows an example sentence from a PubMed article in the Gene Phenotype Relation (PGR) dataset~\cite{sousa2019silver}, which describes the application domain of the present work as well.

% \citet{sousa2019silver} proposed to extract the relations between genes and phenotypes from sentential context as shown in Table \ref{tab:pgr_example_intro}. 
Previous research has extensively investigated relation extraction at both  sentence~\cite{zeng-etal-2015-distant,dos-santos-etal-2015-classifying,sousa2019silver} and document~\cite{yao-etal-2019-docred,quirk-poon-2017-distant} levels. Furthermore, effective graph-based neural network approaches have been developed for various prediction tasks on graphs, including link prediction between given node pairs~\cite{kipf2017semi, hamilton2017inductive,xu2018powerful,velickovicgraph}. Several recent approaches~\cite{lidistance,zhang2018link,alsentzer2020subgraph} illustrated the importance of enhancing graph neural networks using structurally-informed features such as shortest paths, random walks and node position features. 
%\begin{table}[t]\small
%    \centering
%    \begin{tabular}{p{7cm}}
%    \hline
%         \textbf{Sentence}: Our study further emphasizes %that {\underline{NDUFS6}} sequence should be analyzed in %patients presenting with lethal neonatal {\underline %{lactic acidemia}} becuase of isolated complex I %deficiency. \\ %\hline
%         \textbf{Gene}: NDUFS6 \\ %\hline
%         \textbf{Disease}: lactic acidemia \\ %\hline
%         \textbf{Label}: positive relation \\ \hline
%    \end{tabular}
%    \caption{An example showing the report of a positive relation between a gene and a rare disease from the PGR dataset~\cite{sousa2019silver}.}
%    \label{tab:pgr_example_intro}
%\end{table}

In this work, we develop a graph neural network titled \textbf{G}raph \textbf{T}ext \textbf{N}eural \textbf{N}etwork (GTNN) that employs structurally-informed node embeddings as well as textual descriptions of nodes at prediction layer to avoid information loss for relation extraction. GTNN can be trained using a standard approach where data samples are fed to the network in a random order~\cite{hamilton2017inductive}. However, nodes, edges or sub-graphs can significantly vary in their difficulty to learn, owing to frequent substructures, complicated topology and indistinct patterns in graph data. We tackle these challenges by presenting a generic and trend-aware curriculum learning approach that incorporates {\em sample-level} loss trajectories (trends) to better discriminate easier from harder samples and schedule them for training  graph neural networks.

% Trend-SL is built on the top of existing curriculum learning framework SuperLoss (SL)\cite{castells2020superloss} and is over arching which makes the SL a special case of Trend-SL. Our model, GTNN + Trend-SL, takes into account the information of the recent history of the loss for each instance to decide the difficulty threshold at \textit{instance-level} instead of threshold at \textit{batch-level}.

The contributions of this paper are:
(a): a graph neural network that effectively integrates textual data and graph structure for relation extraction, illustrating the importance of {\em direct} use of text embeddings at prediction layer to avoid information loss in the iterative process of learning node embeddings for graph data; and 
(b): a novel curriculum learning approach that incorporates loss trends at sample-level to discover effective curricula for training graph neural networks.

We conduct extensive experiments on real world datasets in both general and specific domains, and compare our model against a range of existing approaches including the state-of-the-art models for relation extraction. Experimental results demonstrate the effectiveness of the proposed approach; the model achieves an average of 8.6 points improvement in F1 score against the best-performing graph neural network baseline that does not directly use text embeddings at its prediction layer. The proposed curriculum learning approach further improves this performance by 0.7 points, resulting in an average F1 score of 89.9 on our three datasets. We conduct extensive experiments to shed light on the improved performance of the model. Code and data are available at \url{https://clu.cs.uml.edu/tools.html}.

\section{Method}\label{sec:model}

\begin{figure*}
    \centering
    \includegraphics[scale=0.33]{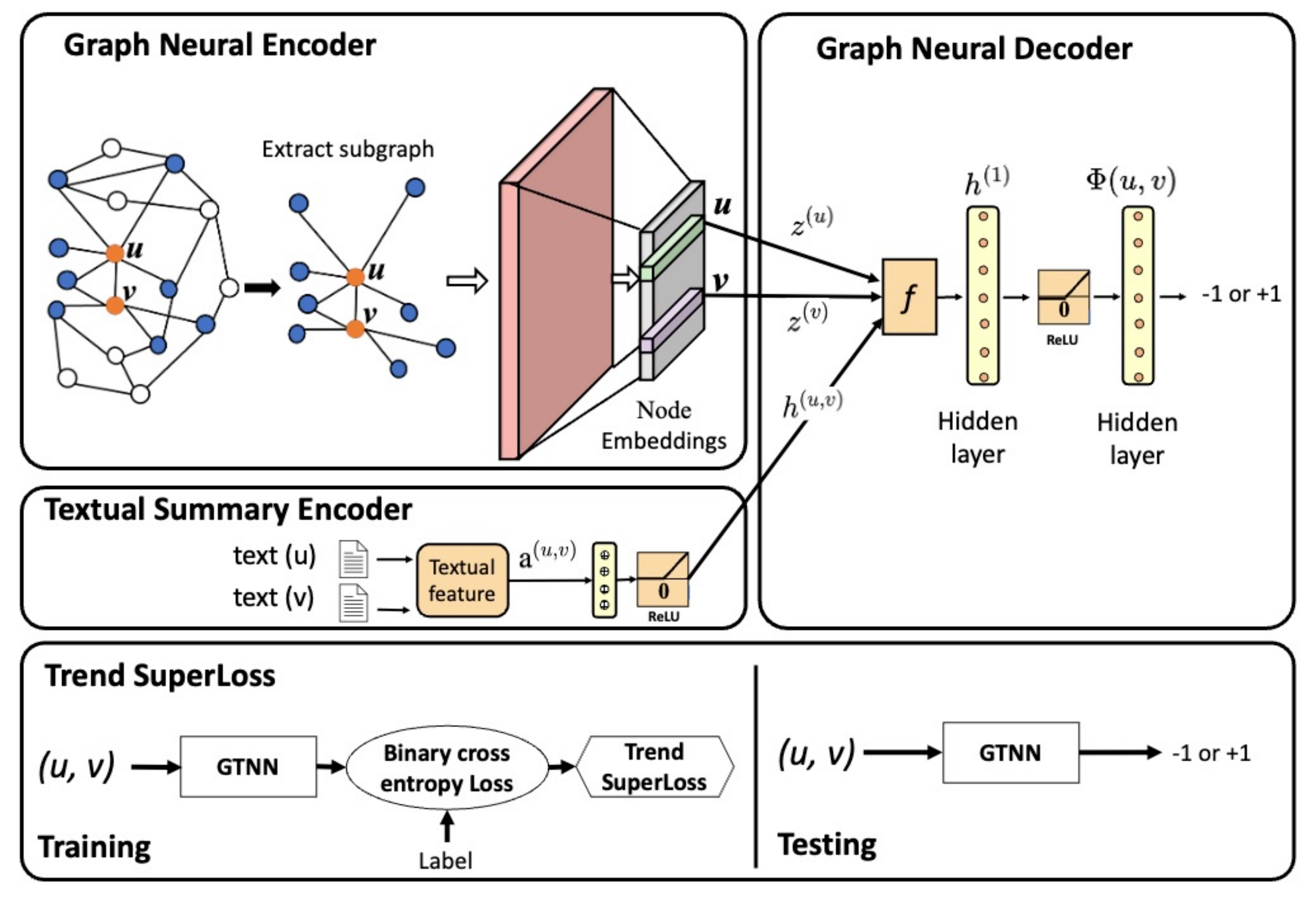}
    \caption{The architecture of the proposed graph text neural network (GTNN) model with Trend-SL curriculum learning approach. The proposed model consists of an encoder-decoder component that determines relations between given node pairs. The graph neural encoder takes as input features from textual descriptions of nodes and sub-graph extracted for a given node pair to create node embeddings. The resulting embeddings in conjunction with additional text features are {\em directly} used by the decoder to predict links between given entity pairs. The resulting loss is given as an input to our Trend-SL approach to dynamically learn a curriculum during training.}
\label{architecture}
\end{figure*}

% We describe our GNN model to incorporate textual features and the curriculum based technique to automatically identify easy and hard examples while training. 

% \subsection{Problem Definition}
Consider an undirected graph $G$ = ($\mathcal{V}$, $\mathcal{E}$) where $\mathcal{V}$ and $\mathcal{E}$ are nodes and edges respectively, and nodes carry text summaries as their descriptions. Edges in the graph indicate ``relations'' between their end points, e.g., causal relations between genes and diseases, or links between concepts in an encyclopedia. 
% Nodes which are not connected by the edges indicates no relation exits between the nodes. 
Our goal is to predict relations/links between given node pairs in $G$.

\subsection {Graph Text Neural Network}
We present the Graph Text Neural Network (GTNN) model which directly operates on $G$ and textual descriptions of its nodes. 
% In what follows, we describe the process of obtaining text embeddings as initial node features, extracting additional text features, graph encoder and decoder models, and well as our curriculum learning training paradigm.   
Figure \ref{architecture} shows the architecture of GTNN, which we describe below.

\subsubsection{Graph Encoder}
Given $G$ and its initial text embeddings, $\bx_i$ for each node $i$, we apply a graph encoder~\cite{hamilton2017inductive} to generate a $d$-dimensional embedding for each node by iteratively aggregating the current embeddings of the node and its $t$-hop neighbours through the {\tt sigmod} function denoted by $g$:
%\sigma
\begin{eqnarray}
\bh_{i}^{(t+1)}=g\Big(\bW_{1}\bh_{i}^{(t)}+
\bW_{2}(\frac{1}{|\mathcal{N}_{i}|}{\displaystyle \sum_{j\in \mathcal{N}_{i}}}\bh_{j}^{(t)})\Big),
\label{eq:graph_equation}
\end{eqnarray}
where ${\bh_i}^{(t)}$ is the embedding of node \textit{i} at the $t^{th}$ layer of the encoder and is initialized by $\bx_i$, i.e., ${\bh_i}^{(0)} = \bx_i,\forall i$, and $\mathcal{N}_i$ is the set of neighbors of node \textit{i} aggregated through a mean operation. ${\bW}_1$ and ${\bW}_2$ are parameter matrices to learn during training. Equation (\ref{eq:graph_equation}), applied iteratively, generates node embeddings $\bz_i = {\bh_i}^{(t+1)}\in\mathbb{R}^d$.

\subsubsection{Additional Text Features}\label{additional_features}
In addition to the representations obtained from the graph encoder, we use additional features from text data to better learn the relations between entities. Here, we consider three types of features: 
(a) relevance score between the descriptions of node pairs obtained from information retrieval (IR) algorithms; we use BM-25~\cite{robertson1995okapi}, classic TF/IDF~\cite{jones1972statistical}, as well as DFR-H~and~DFR-Z~\cite{amati2002probabilistic} models. These IR models capture lexical similarities and relevance between node pairs through different approaches;
(b): we also use the initial text embeddings of nodes ($\bx_i, \forall i$) as additional features because the direct uses of these embeddings at prediction layer can avoid information loss in the iterative process of learning node embeddings for graph data; and 
(c): if there exist other text information for a given node pair, e.g., a sentence mentioning the node pair as in Figure~\ref{fig:pgr_example_intro}, we use the embeddings of such information as additional features.

\subsubsection{Graph Text Decoder}
For a given node pair ($u$,$v$), 
% we have two pieces of information: node representations from the graph encoder as well as additional features obtained from the textual descriptions of the pair. 
we combined representation of their additional features using a single hidden layer neural network as follows:
\begin{equation}
\bh_{uv} = {\tt ReLU}\big(\bW^e\ba_{uv}+\bb^e\big),
\label{eq:additional_feature_hidden_layer}    
\end{equation}
where $\ba$ is obtained by concatenating the additional feature vectors of $u$ and $v$.
We combine $\bh_{uv}$ with node representations, $\bz_u$ and $\bz_v$, and pass them to a two layer decoder to predict their relations:
\begin{eqnarray}
\bh  = {\tt ReLU} \Big(\bW^{last} f(\bh_{uv},\bz_u,\bz_v) +\bb^{last}\Big), \\ \nonumber
p(u,v) = g\left(\bW^{output} \bh +\bb^{output}\right),
\label{similirity}
\end{eqnarray}
where $f$ is a fusion operator, $g$ is the {\tt sigmod} function, and $p(u,v)$ indicates the probability of an edge between nodes $u$ and $v$. Flattened outer product, inner product, concatenation and 1-D convolution can be used as the fusion operator~\cite{amiri-etal-2021-attentive}. In our experiments, we obtained better performance using outer product, perhaps due to its better encoding of feature interactions: 
\begin{equation}
    f(\bh_{uv},\bz_u,\bz_v) = \bh_{uv} \otimes [\bz_u;\bz_v].
\end{equation}

\subsection{Generic Trend-aware Curricula}
% The above graph neural network approach can be trained using a binary cross entropy loss function~\cite{hamilton2017inductive}. 
% \begin{equation}
% J(u,v) & = &y_{uv}log\left(p(u,v\right)\label{Loss_of_a_pair}+\left(1-y_{uv}\right)log\left(1-p(u,v)\right)\nonumber
% \end{equation}
% where $J(u,v)$ is the loss of predicting an edge for the $(u,v)$ pair,
% $y_{uv}\in\{0,1\}$ is the ground truth label for the relation between $u$ and $v$, and $p(u,v)$ is the probability of the edge $(u,v)$.
%
Graph neural networks are often trained using the standard or ``rote'' approach where samples are fed to the network in a random order for training~\cite{hamilton2017inductive}. However, edges (and other entities in graphs such as nodes and sub-graphs) can vary significantly in their classification difficulty, and therefore we argue that graph neural networks can benefit from a curriculum for training. Recent work by~\citet{castells2020superloss} described a generic loss function called SuperLoss (SL) which can be added on top of any target-task loss function to dynamically weight training samples according to their difficulty for the model. Specifically, it uses a {\em global} difficulty threshold ($\tau$), determined by the exponential moving average of all sample losses, and considers samples with an instantaneous loss smaller than $\tau$ as easy and the rest as hard. Similar to the commonly-used easy-to-hard transition curricula, such as those in~\cite{bengio2009curriculum}~and~\cite{kumar2010self}, the model initially assigns higher weights to easier samples, thereby allowing back-propagation to initially focus more on easier samples than harder ones. 

However, SL does not take into account the trend of instantaneous losses at sample-level, which can  
(a): improve the difficulty estimations of the model by making them {\em local}, {\em sample dependent} and potentially more {\em precise}, and 
(b): enable the model to distinguish samples with similar losses based on their known loss trajectories. For example, consider an easy sample with a rising loss trend which is about to become a hard sample versus another easy sample with the same instantaneous loss but a falling loss trend which is about to become further easier for the model. Trend information allows distinguishing such examples. 
% In addition, we observe that the estimated difficulty of a considerable fraction of samples with a falling or rising loss trend will be inverted by SL during training (see Section~\ref{sec:inversions}), which indicates the importance of trend information during training.  

% 21.2\% to 50.0\% of hard samples that have a falling loss trend will become easy in the next iteration; similarly 1.3\% to 11.1\% of easy samples that have a rising loss trend will become hard in the next iteration. 

% auc for easy with rising trend converting to hard : 24.87
% auc for hard with falling trend. converting to easy : 4.51

\begin{figure}
    \centering
    \includegraphics[scale=0.22]{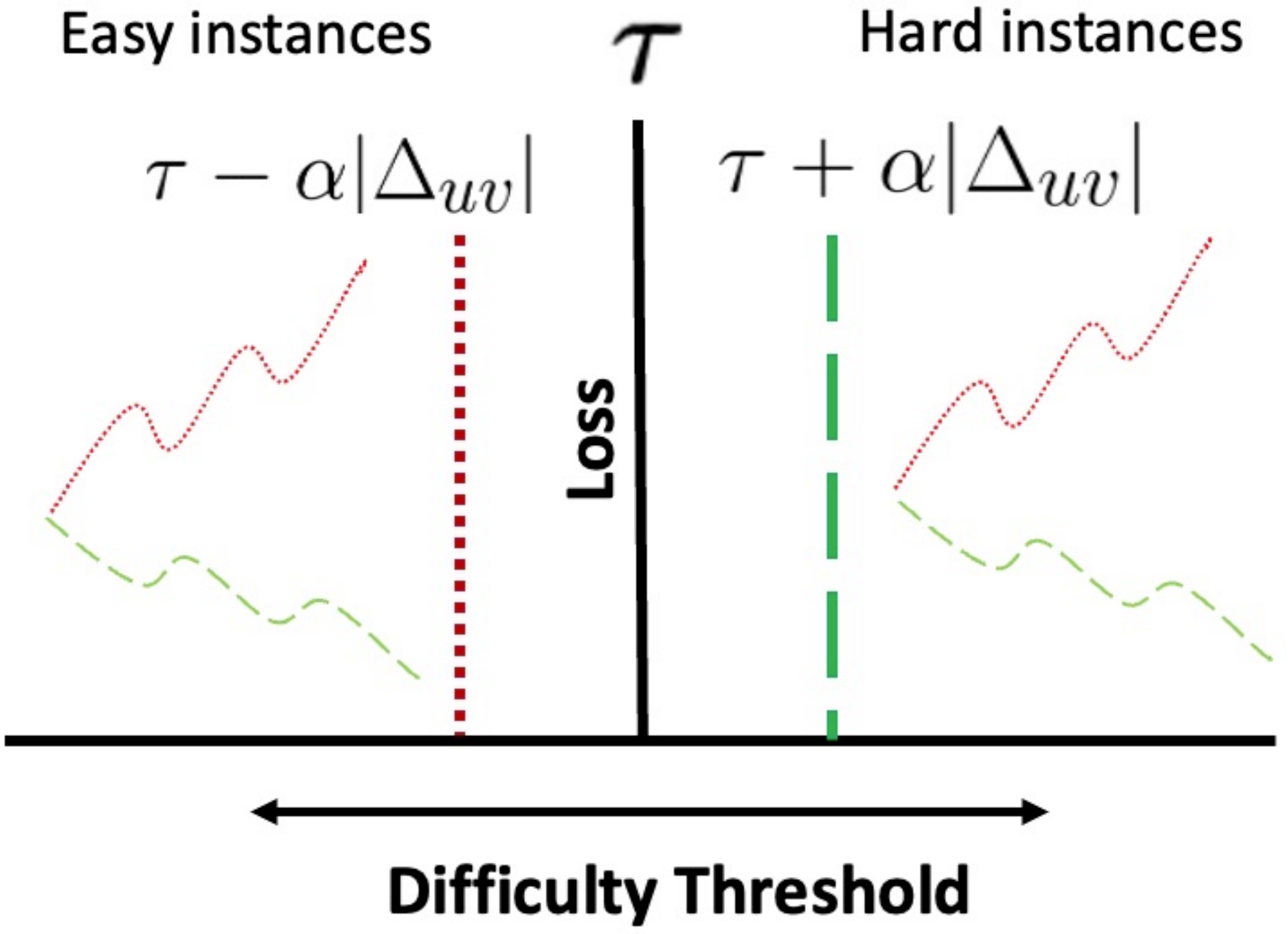}
    \caption{Difficulty dynamics in Trend-SL. $\tau$ is the fixed difficulty threshold of SL, which can be thought of as a global difficulty metric to separate easy and hard samples. Dotted (red) and dashed (green) trend lines indicate four samples with rising and falling loss trends respectively. Trend-SL uses trend dynamics to shift the difficulty boundaries and adjust global difficulty using local sample-level loss trends. The vertical dashed and dotted lines show updated sample-specific difficulty thresholds for easy and hard samples respectively. }
    \label{fig:sl_vs_tsl}
\end{figure}

The above observations inspire our work to utilize trend information in our curriculum learning framework, called Trend-SL. The model uses loss information from the local time window before each iteration to capture a form of momentum of loss in terms of rising or falling trends and determine individual sample weights as follows:
\vspace{-0.2cm}

{\small{
\begin{eqnarray}\label{eq:trend_sl}
TrendSL_{\lambda,\alpha}(l_{uv}) = \arg\min_{\sigma_{uv}} \big(l_{uv}- (\tau-\alpha\Delta_{uv}) \big) \\\nonumber
\times \sigma_{uv} +\lambda(\log\sigma_{uv})^{2},
\end{eqnarray}}}
where $\sigma_{uv}$ is the latent weight for the training sample $(u,v)$ , $l_{uv}$ is the target-task loss (binary cross-entropy in our experiments) for $(u,v)$ at current iteration,
$\tau$ is the batch-level global difficulty threshold determined by the exponential moving average of sample losses~\cite{castells2020superloss}, %and can be thought of as a global difficulty threshold, 
and $\Delta\in[-1,1]$ is the trend indicator quantified by the normalized sample-level loss trend weighted by $\alpha\in[0,1]$; our approach reduces to SL with $\alpha=0$.
%
% SuperLoss sets the threshold at batch level containing instances of various difficulty levels hence, it does not appropriately estimate the threshold for each instances. In contrast to SuperLoss, Trend-SL considers the batch threshold and adjusts the threshold at instance level based on the trend information of the instance.
%
$\Delta$ captures the trend in the instantaneous losses of samples over recent $k$ iterations, effectively utilizing local sample-level information to determine difficulty. There are various techniques for fitting trends to time series data~\cite{bianchi1999comparison}. We use differences between consecutive losses to determine the trend for each sample:
\vspace{-0.4cm}
\begin{equation}\label{eq:trend_sl_delta}
\Delta_{uv} = \stackrel[j=i-k+2]{i}{\sum} (l_{uv}^{j}-l_{uv}^{j-1}) / \stackrel[j=i-k+2]{i}{\sum}\mid l_{uv}^{j}-l_{uv}^{j-1}\mid,
\end{equation}
where $i$ is the current iteration, $l_.^j$ indicates loss at iteration $j$ and $k$ controls the number of previous losses to consider.
As Figure~\ref{fig:sl_vs_tsl} illustrates, Trend-SL increases the difficulty threshold for samples with falling loss trends (negative $\Delta$s), becoming more flexible in increasing the weights of such samples by allowing greater instantaneous losses. On the other hand, it becomes more conservative in weighting samples with rising trends (positive $\Delta$s) by reducing the difficulty threshold. 

Finally, we note that the weight $\sigma_{uv}$ in (\ref{eq:trend_sl}) can be computed as follows, where $W$ is the Lambert W function~\citep{euler1783serie}; see details in the supplementary materials in~\citep{castells2020superloss}: 
\begin{eqnarray}
\sigma_{uv}^* & = & \exp{\Big(-W \big(\frac{1}{2}\max (-\frac{2}{e}, \beta)\big)\Big)},\\ 
\beta & = & \frac{l_{uv}-\left(\tau-\alpha\Delta_{uv}\right)}{\lambda}.
\end{eqnarray}
% {\color{red}We optimize Eq. \ref{eq:trend_sl} across all the training examples to train our model.}
% Finally, the total loss, $J_{total}$, can be computed using positive and negative training edges as follows:
% \begin{eqnarray}
% J_{total} = \sum \sigma_{uv} \times J\left(u,v\right), \forall (u,v)\in\emph{training}.
% \label{total_loss}
% \end{eqnarray}

% \cut{\begin{table}[t]\small
%     \centering
%     \begin{tabular}{|l|l|}\hline  
%       {\bf Metric} & {\bf Value} \\\hline 
%          \# Gene nodes &  4,274\\\hline
%          \# Disease nodes & 6,143\\\hline
%          \# Disease type nodes & 472\\\hline
%          \# Phenotype nodes & 9,603\\\hline
%          \# Nodes in LCC & 20,264 \\\hline
%          \# Edges & 964,222 \\\hline
%              \hspace{5pt} $\rightarrow$ Gene-Disease & 6,284 \\\hline
%              \hspace{5pt} $\rightarrow$ Gene-Phenotype & 595,296 \\\hline
%              \hspace{5pt} $\rightarrow$ Disease-Disease type & 3,912 \\\hline
%              \hspace{5pt} $\rightarrow$ Disease-Phenotype & 358,730 \\\hline
%          \# Edges in LCC & 964,087 \\\hline
%          \# Connected components & 94 \\\hline
%          Average node degree & 94.11 \\\hline
%              \hspace{5pt} $\rightarrow$ Genes & 140.75 \\\hline
%              \hspace{5pt} $\rightarrow$ Diseases & 60.04 \\\hline
%              \hspace{5pt} $\rightarrow$ Disease types & 8.28 \\\hline
%              \hspace{5pt} $\rightarrow$ Phenotypes & 99.35 \\\hline
%          Graph density & 0.004592 \\\hline
%     \end{tabular}
%     \caption{Statistics of Gene, Disease, Phenotype Relation (\gdpr) dataset.}

%     \label{tab:omim_dataset}
% \end{table}
% }

\section{Experiments} \label{sec:experiments}
% We pre-process and compute textual features, and use a targeted sampling approach to create negative examples for training and evaluation purpose, see ``Datasets'' below. We evaluate all approaches on three datasets (Table \ref{tab:stats}) (1) GDPR, (2) PGR, and (3) Wikipedia (Chameleon).

\subsection{Datasets}
\paragraph{Gene, Disease, Phenotype Relation ({\gdpr})} dataset contains textual descriptions for genes, diseases and phenotypes (symptoms) as well as their relations, and is obtained by combining two freely available datasets: Online Mendelian Inheritance in Man (OMIM)~\cite{amberger2019omim} and Human Phenotype Ontology (HPO)~\cite{hpo}. OMIM is the primary repository of curated information on the causal relations between genes and rare diseases, and HPO provides mappings of phenotypes to genes/diseases in the OMIM.\footnote{A gene can cause one or more diseases and a disease can have several disease types. As a pre-processing step, we remove isolated nodes from the dataset and explicit mentions of relations between entities from summaries.}
%and remove mentions of genes and phenotypes from each other's description.
% We use these mappings and the textual descriptions of entities in these datasets to compile \gdpr.
%% See the disease Leukemia: \url{https://omim.org/entry/608232}.
% \textbf{Disease-Gene Data for Differential Diagnosis}
We introduce a challenging experimental setup based on the task of {\em differential diagnosis}~\cite{raftery2014churchill} using \gdpr, where competing models should distinguish relevant diseases to a gene from irrelevant ones that present {\em similar} clinical features, making the task more difficult because of high textual and structural similarity between relevant and irrelevant diseases. For example, diseases {\it 3-methylglutaconic type I}, {\it Barth syndrome} and {\it 3-methylglutaconic type III} are of the same disease type and have high lexical similarity in their descriptions, but they are not related to the same genes.
We include such harder negative gene-disease pairs by sampling genes from those that are linked to diseases that share the same disease type with a target disease, but are not linked to the target disease. We also include an equal number of randomly sampled negative pairs to this set.

\paragraph{Gene Phenotype Relation (\pgr)}~\cite{sousa2019silver} is created from PubMed articles and contains sentences describing relations between given genes and phenotypes ( Figure~\ref{fig:pgr_example_intro}). %in the Introduction. 
% \pgr~and~\gdpr~have overlap in their gene and phenotype sets. 
We only include data points with available text descriptions for their genes and phenotypes. For fair comparison, we apply the best model from~\cite{sousa2019silver} to this dataset.
% \footnote{We note that we applied their model to the full \pgr~dataset and were able to reproduce similar score as mentioned in~\cite{sousa2019silver}.}

\paragraph{Wikipedia}~\cite{rozemberczki2021multi} is on the topic of the old world lizards Chameleons with 202 species. In this dataset, nodes represent pages and edges indicate mutual links between them. Each page has an informative set of nouns, which we use as additional features. We note that this dataset contains only these noun features but not the original text, which is required by our text only models.

% As there is no explicit text associated with each page, we have used the given noun features as node features  and as an additional textual features in our task. 
% Given the large vocabulary size of these nouns, we reduce the dimensionality of these features through PCA~\cite{wold1987principal} to use in graph neural networks. 

% Table generated by Excel2LaTeX from sheet 'Sheet4'

    % \# Nodes & 18,330 & 20,489 & 2,227 \\ %\hline
    % \# Edges & 365,014 & 605,492 & 31,421 \\ %\hline
    % \# {\color{red}{Annotated}} Edges & 37,693 & 3,049 & 188,526 \\ %\hline
    % \hspace{5pt} $\rightarrow$ \# pos. Edges & 6,284 & 1,445  & 31,421 \\ %\hline
    % \hspace{5pt} $\rightarrow$ \# neg. Edges & 31,409 & 1,604  & 157,105 \\ \hline
    
\begin{table}[t]\small
  \centering
    \begin{tabular}{llll}
    \hline
    \textbf{Metric} & \textbf{GDPR} & \textbf{PGR} & \textbf{Wikipedia} \\ \hline
    \# Nodes & 18.3K & 20.4K & 2.2K \\ %\hline
    \# Edges & 365.0K & 605.4K & 31.4K \\ %\hline
    \# Sampled Edges & 37.6K & 3.0K & 188.5K \\ %\hline
    \hspace{5pt} $\rightarrow$ \# pos. Edges & 6.2K & 1.4K  & 31.4K \\ %\hline
    \hspace{5pt} $\rightarrow$ \# neg. Edges & 31.4K & 1.6K  & 157.1K \\ \hline
    \end{tabular}
  \caption{Statistics of the three datasets. Sampled edges are used to create training, validation and test sets. All models take the entire graph as input.}
  \label{tab:stats}
\end{table}

Table~\ref{tab:stats} shows statistics of these datasets. In case of \gdpr{} 
and \wikipedia{}, we create five negative examples for every positive pair.
We divide these pairs into $80$\%, $10$\% and $10$\% as training, validation and test splits respectively. The data splits for \pgr{} is the same as the original dataset, except that we discard data points (node pairs) that do not have text descriptions.
% \footnote{
% To prevent models from memorizing node pair labels, 
% instead of finding their relation to the actual task, 
% We include harder negative examples, e.g., by sampling negative genes from those that are linked to diseases that share the same disease type with a target disease, but are not linked to the target disease. We also include an equal number of randomly sampled negative examples to this set.}  

% \begin{table}[h]
%     \centering
%     \begin{tabular}{|p{7cm}|}
%     \hline
%          \textbf{Sentence}: Our study further emphasizes that {\underline{NDUFS6}} sequence should be analyzed in patients presenting with lethal neonatal {\underline {lactic acidemia}} due to isolated complex I deficiency. \\ \hline
%          \textbf{Gene}: NDUFS6 \\ \hline
%          \textbf{Phenotype}: lactic acidemia \\ \hline
%          \textbf{Ground Truth}: Causal \\ \hline
         
%     \end{tabular}
%     \caption{An example from \pgr{} dataset. Here, we map gene (NDUFS6) and phenotype (lactic acidemia) to the \gdpr dataset to learn node embeddings of the gene and phenotype and use it for classification along with sentence embedding as an additional feature.}
%     \label{tab:pgr_example}
% \end{table}

\subsection{Baselines} \label{sec:baselines}
We use the following baselines:

\begin{itemize}[leftmargin=*]
    \setlength{\itemsep}{0pt}
    \setlength{\parsep}{0pt}
    
    \item \textbf{Co-occurrence} labels a test pair as positive if both entities occur together in the input text.
    % As every pair is classified as positive relation by this baseline, the recall of this model is always $1$. 
    
    \item \textbf{Relevance Score} uses scores from IR models (Section~\ref{additional_features}) as features of a logistic classifier.
    % to predict the relations between input pairs.  
    
    \item \textbf{Doc2Vec}~\cite{le2014distributed} uses domain-specific text embeddings obtained from Doc2Vec
    % from GenSim~\cite{rehurek2011gensim} 
    as features of a logistic classifier.
    
    % to initialize node representation for \gtnn

    \item \textbf{BioBERT}~\cite{lee2020biobert,devlin2018bert} is a BERT model pre-trained on PubMed articles. BioBERT is most appropriate for relation extraction on both \gdpr~and~\pgr~datasets as they are also developed based on PubMed articles. It is the current state-of-the-art model on \pgr~\cite{sousa2019silver}. We also include a version of BioBERT that uses graph information by concatenating the representation of each given pair with the average embedding of its neighbors.
    
    \item \textbf{Graph Convolutional Network} (GCN)~\cite{kipf2017semi} is an efficient and scalable approach based on convolution neural networks which directly operates on graphs.
    
    \item \textbf{Graph Attention Network} (GAT)~\cite{velickovicgraph} extends GCN by employing self-attention layers to identify informative neighbors while aggregating their information, effectively prioritizing important neighbors for target tasks. 
    
    \item \textbf{GraphSAGE}~\cite{hamilton2017inductive} is an inductive framework which aggregates node features and network structure to generate node embeddings, see (\ref{eq:graph_equation}). It uses both text and graph information. We use Doc2Vec~\cite{le2014distributed} embeddings to initialize node features of GraphSAGE, as they led to better performance than other embeddings in our experiments. 
    
    \item \textbf{Graph Isomorphism Network} (GIN)~\cite{xu2018powerful} identifies the graph structures that are not distinguishable by the variants of graph neural networks like GCN and GraphSAGE. Compared to GraphSAGE and GCN, GIN uses extra learnable parameters during sum aggregation and uses MLP encoding.
    
    %\item \textbf{Distance Encoding} (DE)~\cite{lidistance} uses distance based features (shortest path and landing probabilities in random walk of different lengths) to create better node representations. These features are used as controllers of message aggregation. Calculating the landing probability on our datasets is extremely time consuming due to the specific model requirements and the large sizes of our datasets, and we only use the shortest path in this approach as distance features. 
    
    \item \textbf{CurGraph}~\cite{wang2021curgraph} is a curriculum learning framework for graphs that computes difficulty scores based on the intra- and inter-class distributions of embeddings and develops a smooth-step function to gradually include harder samples in training. We report the results of our implementation of this approach.   
    
    \item \textbf{SuperLoss} (SL)~\cite{castells2020superloss} is a generic curriculum learning approach that dynamically learns a curriculum from model behavior. It uses a fixed difficulty threshold at batch level, determined by the exponential moving average of all sample losses, to assign higher weights to easier samples than harder ones.

\end{itemize}

We compare these baselines against \textbf{GTNN} and \textbf{Trend-SL}, described in Section~\ref{sec:model}.

\begin{table*}[htbp]\small
  \centering
    \begin{tabular}{l p{3cm} l l l  l l l  l l l  c}
     \hline
        \textbf{Modality}  & \textbf{Model} & \multicolumn{3}{c}{\textbf{GDPR}} & \multicolumn{3}{c}{\textbf{PGR}} & \multicolumn{3}{c}{\textbf{Wikipedia}} &  \\
          %\hline
          &       & \textbf{P}     & \textbf{R}     & \textbf{F1}    & \textbf{P}     & \textbf{R}     & \textbf{F1}    & \textbf{P}     & \textbf{R}     & \textbf{F1}    & \multicolumn{1}{c}{\textbf{avg F1}} \\
          \hline
        %   \multicolumn{12}{|c|}{\textbf{Without curriculum Learning}} \\
        %   \hline
    -      & Co-occurance & 16.7  & 100   & 28.6  & 47.5  & 100   & 64.4  & 16.7  & 100   & 28.6  & 40.5 \\
    T     & Relevance Score  & 59.2 & 83.4 & 69.2 & 75.6 & 64    & 69.1 &   -    &  -     & -  & 69.2 \\
    %T     & BioBERT & N/A   & N/A   & N/A   & 81.8  & 72    & 76.6  & N/A   & N/A   & N/A   & N/A \\
    T     & BioBERT (node pairs) & 20.3 & 55.6 & 29.7 & 84.9 & 74.7 & 79.4 &  -   & -  & -   & 54.6\\
    T     & BioBERT (neighbors) & 21.1 & 57.4 & 30.9 & 74.0 & 76.0    & 75.0   & -   & -   & -   & 53.0 \\
    T     & Doc2vec (node pairs) & 19.8 & 45.0 & 27.5  & 80.5 & 82.7 & 81.6 & -   & -   & -   & 54.6 \\
    T    & Doc2vec (neighbors) & 20.6 & 51.9 & 29.5 & 83.1  & 78.7 & 80.8 & -   & -   & -   & 55.2\\ \hline
    G     & GCN   & 34.2 & 44.5 & 38.6 & 61.1 & 79.5 & 68.6 & 72.8 & 89.7 & 80.3 & 62.5 \\
    G     & GAT   & 23.7 & 50.3 & 31.7 & 75.8 & 91.1 & 82.5  & 78.2 & 86.7 & 82.2 & 65.5 \\
    G     & GIN   & 21.8 & 48.1 & 29.8 & 54.2 & 88.1  & 67.0 & 76.4 & 77.2  & 76.1 & 57.6 \\
    G     & GraphSAGE (random) & 17.2 & 90.4 & 28.5 & 84.8  & 79.2 & 81.8 & 57.9 & 82.28  & 67.9  &  59.4 \\ \hline
    G,T   & GraphSAGE (Doc2Vec) & 54.0 & 79.2 & 64.1 & 91.8 & 90.2  & 91.0 & 81.5 & 93.0 & 86.6 & 80.6 \\
    %G,D   & DE-GNN & 79.6 & 76.4 & 77.8 & 85.9 & 98. & 91.8 &       &       &       & 84.8 \\
    G,T   & GTNN  & 78.0 & 87.9 & \textbf{82.6} & 93.6  & 93.2 & \textbf{93.4} & 87.9 & 95.4 & \textbf{91.5} & \textbf{89.2} \\
    \hline
    % \multicolumn{12}{|c|}{\textbf{With Curriculum Learning}} \\
    % \hline
    % G     & CurGraph & 19.3 & 79.9 & 30.0 & 57.1 & 94.6 & 70.6 &   19.6 & 88.9 & 30.6  & 33.6 \\
    % G,T   & GTNN + SL & 80.7  & 86.5 & 83.5 & 93.4 & 94.6 & 94.0 & 90.0 & 94.0 & \textbf{92.0} & 89.8 \\
    % G,T   & GTNN + Trend-SL & \textbf{83.0} & 85.7 & \textbf{84.3} & \textbf{95.5} & 92.9  & \textbf{94.2} & \textbf{89.8} & 92.8 & 91.3 & \textbf{89.9} \\
    % \hline
    \end{tabular}
  \caption{Performance of different models on \gdpr{}, \pgr{}, and \wikipedia{} datasets. 
%   For a fair comparison for bench-marking, we evaluated gene-disease relations present in the \gdpr graph using existing \pgr dataset as it has been used before and gene-phenotypes pairs present in \pgr  overlaps  with \gdpr. 
  Here, (T) indicates ``Text only", (G) indicates ``Graph only", (G,T) indicates combination of both.
  %, and (G,D) indicates use of the Distance metric as an additional features. 
  Note that the \wikipedia{} dataset contains only noun features but not the original text, which is required by the text only models.}.
  \label{tab:performance}
\end{table*}%

% Table generated by Excel2LaTeX from sheet 'Sheet2'
\begin{table}[htbp]\footnotesize
  \centering
  
    \begin{tabular}{ l c c c c }
     \hline
    \textbf{Model} & \textbf{GDPR} & \textbf{PGR} & \textbf{Wikipedia}  & \textbf{avg F1}\\
    \hline
    \textbf{GTNN} & 82.6 & 93.4 & 91.5 & 89.2 \\
    \textbf{CurGraph} & 75.9 & 85.1 &  80.3  & 80.3 \\
    \textbf{SL} & 83.5  & 94.0  &\textbf{ 92.0}  & 89.8 \\
    \textbf{Trend-SL} & \textbf{84.3}  & \textbf{94.2 } & 91.3  & \textbf{89.9 }\\
    \hline
    \end{tabular}%
    \caption{Performance of curriculum models on \gdpr{}, \pgr{}, and \wikipedia{} datasets. The base model for all curriculum learning approaches is GTNN, see the last row in Table~\ref{tab:performance}.}
  \label{tab:curricula}%
\end{table}%

\subsection{Settings}

We reproduce the results reported in~\cite{sousa2019silver} using BioBERT and therefore follow the same settings on the \pgr~dataset. 
% The features from BioBERT are of dimension $768$ which is used to train a Logistic Regression classifier with {\tt L2} penalty. 
% We also train a Logistic Regression classifier for the Relevance Score baseline for all datasets.
% We assume that every node in the input graph carries a textual summary as its definition. We encode the semantics of the textual summaries using neural network based model depending on the domain. 
Initial domain-specific node embeddings are obtained using Doc2Vec~\cite{le2014distributed} or Bio-BERT~\cite{lee2020biobert}. In case of Bio-BERT, since nodes carry long descriptions, we first generate sentence level embeddings and use their average to represent each node, following~\cite{zhang2019bertscore}. More recent techniques can be used as well~\cite{beltagy2020longformer}.
%For cases Chameleon (Wikipedia) dataset, we used given informative noun features associated with the pages of the Wikipedia articles.
%For \gdpr{} and \pgr{}, these summaries are collected from \omim \cite{amberger2019omim} and \hpo \cite{hpo} datasets.\footnote{Some sections of summaries may contain explicit mentions of relations between entities. As a pre-processing step, we remove all mentions of genes or diseases from summaries.}
%
We consider 1-hop neighbors and set $t=1$ in (\ref{eq:graph_equation}).
% as immediate neighbors are expected to better contribute to relations in the graph. 
To optimize our model, we use the Adam optimizer~\cite{kingma2014adam} and apply hyper-parameter search and tuning for all competing models based on performance on validation data. 
% from $\{0.01, 0.001, 0.0001, 0.00001\}$, 
% employ {\tt L2} regularizer
% from $\{0.01, 0.001, 0.0001\}$ for hyper-parameter search and tuning. 
In (\ref{eq:trend_sl}), we set $\alpha$ from $[0, 1]$ with a step size of 0.1, $\lambda$ from $\{0.1, 0.5, 1.0, 5, 10, 100\}$, and loss window $k$ from $[1,10]$ with a step size of 1. We consider a maximum number of $100$ training iterations with early stopping based on validation data for all models. 
In addition, we evaluate models based on the standard Recall, Precision and F1 score for classification tasks \cite{sklearn_api}. 
% We report F1 scores as an average across 5 runs.
We experiment with five random seeds and report the average results. For all experiments, we use Ubuntu 18.04 with one 40GB A100 Nvidia GPU, 1 TB RAM and 16 TB hard disk space. GPU hours to train our model have been linear to the size of the datasets ranging from 30 min to 5 hours.
We use Precision (P), Recall (R) and F1 score (F1) as evaluation metrics.

\subsection{Results}
Table~\ref{tab:performance} shows the results. We start with text only and graph only baselines followed by baselines that incorporate both data modalities.  

\paragraph{Text models (T):} Comparing all text based model, Relevance Score and Doc2Vec outperform other models. In case of \gdpr, high performance of Relevance Score indicates the ability of unsupervised IR models in finding relevant information in long text descriptions. However, Relevance Score shows poor performance on \pgr{} compared to Doc2Vec, which is better at semantic representation of input data. 
% This is perhaps because, in \pgr{}, phenotypes have very short descriptions, making it hard for IR algorithms to find meaningful signals. 
BioBERT (node pair) obtains higher precision on both datasets and good performance on \pgr.
In addition, the F1 score of the BioBERT model developed in~\cite{sousa2019silver} for \pgr{} is 76.6. 
We note that Doc2Vec obtains better performance than BioBERT, perhaps due to its in-domain pre-training.

\paragraph{Graph models (G):} The results show that GCN and GAT perform better than other competing graph models. We attribute their performance to the use of convolution and attention networks, which effectively prioritize important neighboring nodes with respect to the target tasks.  

\paragraph{Graph models with additional information:} Comparing GraphSAGE (Doc2Vec) and GraphSAGE (random) illustrates the significant effect of initialization with in-domain embeddings. In addition, GTNN outperforms GraphSAGE, resulting in an average of 8.6 points improvement in F1 score. This improvement is because GTNN {\em directly} uses text descriptions at its prediction layer. This information, although available to GraphSAGE as well, can be lost in the iterative process of learning node embeddings through neighbors, see (\ref{eq:graph_equation}). 
%GTNN also outperforms DE-GNN, which indicates the greater importance of text features as apposed to structural/graph features in NLP tasks.  

\paragraph{Training with curricula:} The results in Table~\ref{tab:curricula} show that training GTNN with effective curricula can further improve its performance. We attribute the better performance of Trend-SL compared to SL to the use of trend information, which leads to better curricula. We conduct further analysis on the effect of trend information below. The lower performance of CurGraph could be due to close probability densities that we obtained for samples in our datasets, which do not allow easy and hard samples to be effectively discriminated by CurGraph.

% CurGraph with other baselines in curriculum learning models,  CurGraph shows poor performance over all. This is due to the fact that static curriculum is not as robust  as compared to dynamic curriculum. Overall our model GTNN+Trend-SL performs better than GTNN+SL. This is due to incorporating history information of losses for each instance which is not present in SL. 

\section{Trend Model Introspection}\label{sec:discussion}

We conduct several ablation studies to shed light on the improved performance of Trend-SL. 
%Analysis on the effects of node embedding initialization and additional textual features can be found in supplementary materials.

\subsection{Inversion Analysis}\label{sec:inversions}

\paragraph{Trend-SL results in robust estimation of difficulty:} In curriculum learning, instantaneous sample losses can fluctuate as model trains~\cite{zhou2020curriculum}. These changes result in samples being moved across easy and hard data groups. Let's define an \textit{inversion} as an event where the difficulty group of a sample is inverted in two consecutive epochs (determined by curricula), i.e., an easy sample becomes hard in the next iteration or vice versa. Figure~\ref{fig:flip_analysis} shows the number of inversions in SL and Trend-SL during training. Both models converge on their estimated difficulty classes of samples as training progresses. However, we observe that Trend-SL results in fewer inversions compared to SL, as the area under the curve for Trend-SL is 2.12 compared to 2.15 of SL. 
%is naturally decreasing for both 
Given these results and the performance of Trend-SL on our target tasks, we conjecture that trend information leads to more robust estimation of sample difficulty. 

\begin{figure}[!t]
    \centering
    %0.4
    \includegraphics[scale=0.4]{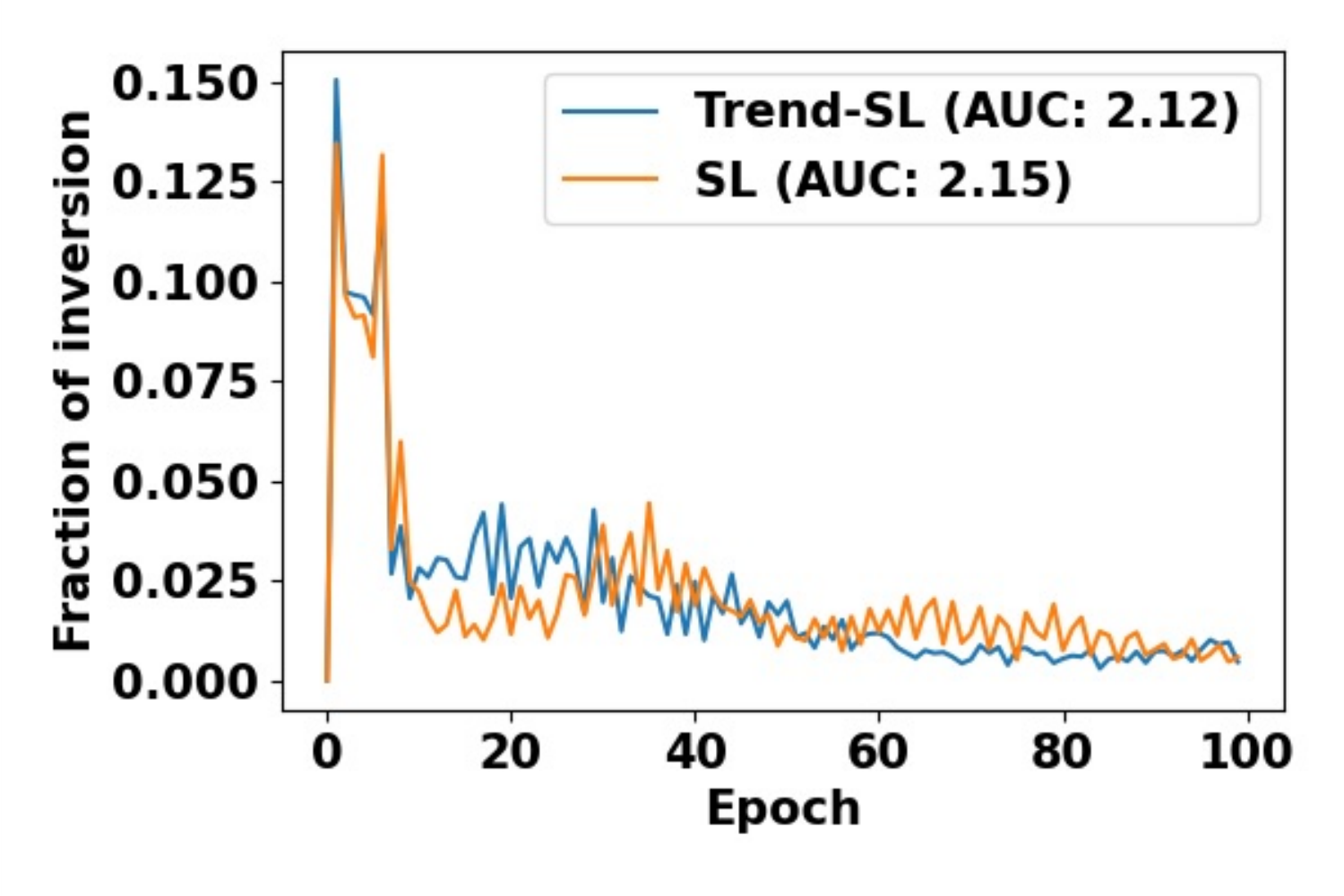}\label{fig:flip_analysis}
    \caption{The fraction of samples with an inverted difficulty group in two consecutive epochs. Both models are converging on the their estimated difficulty classes of samples as training progresses. Trend-SL results in fewer inversions compared to SL; the area under the curve for Trend-SL is 2.12 compared to 2.15 of SL.}
    \label{fig:flip_analysis}
\end{figure} 

\begin{figure}[!t]
    \centering
    \includegraphics[scale=0.55,clip]{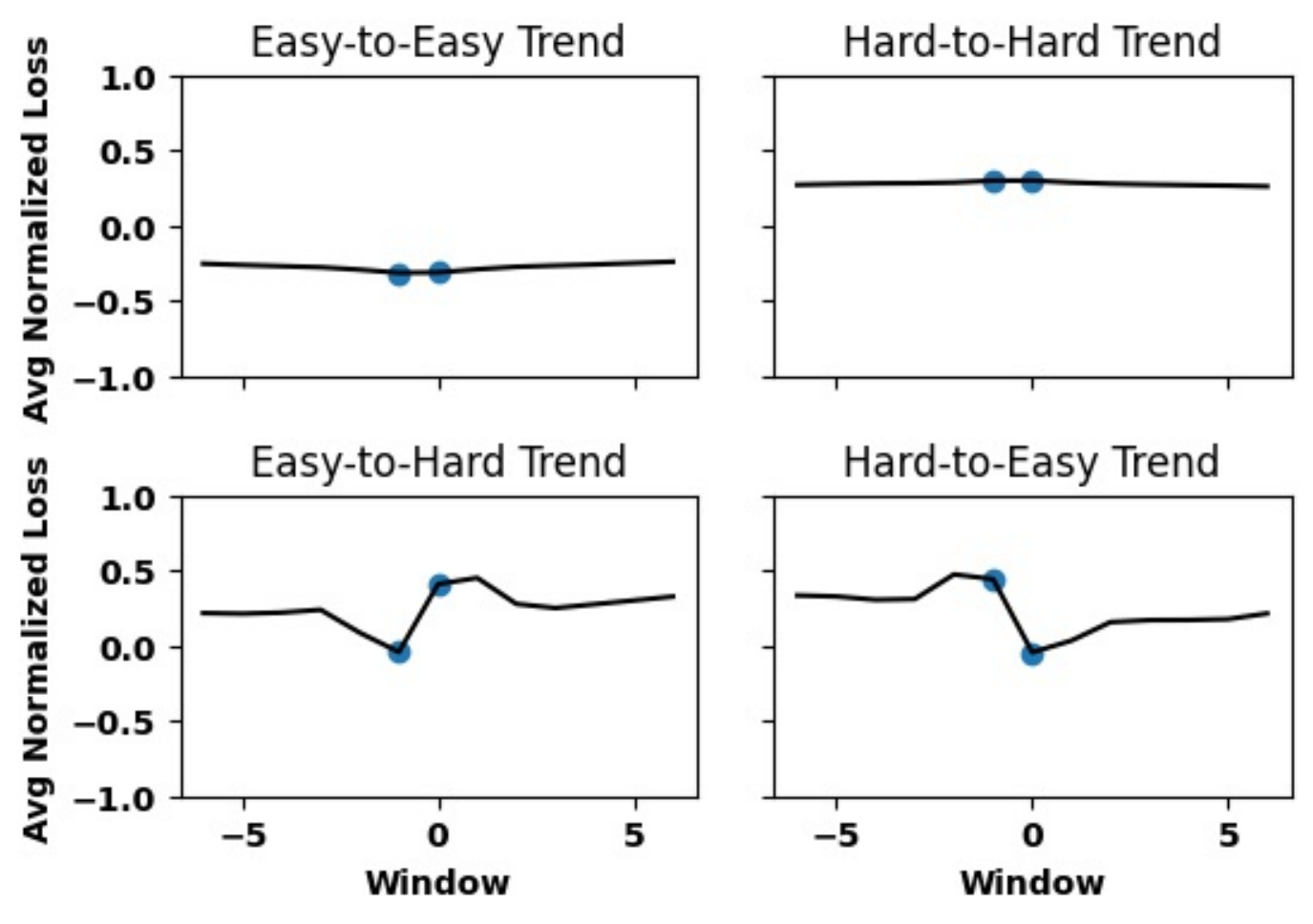}
    \label{fig:trend_window}
    \caption{Transition in sample difficulty determined by SL. 0 on the x-axis denotes any epoch at which an inversion occurs, and the y-axis shows average normalized losses at epochs around the inversion epochs. %Easy-to-Easy and Hard-to-Hard transitions are almost flat lines, while 
    Easy-to-Hard and Hard-to-Easy transitions show sharp and significant increase and decrease in losses respectively.}
    \label{fig:transition}
\end{figure}

\paragraph{Transition patterns at inversion time:}
% We conduct fine-grained analysis on the trend of loss at and around flip time.
% Figure \ref{fig:transition} shows the comparison between the two curricula learned by SL and Trend-SL on \gdpr{} dataset. 
Let epoch $e$ be the epoch at which an inversion occurs. %, i.e., an easy sample becomes hard or vice versa.
Considering SL as the curriculum, Figure~\ref{fig:transition} reports the average normalized loss of samples at their inversion epochs ($e$) and $k$ epochs before and after that. There are some insightful patterns: 
(a): easy-to-easy (E2E) and hard-to-hard (H2H) transitions are almost flat lines, indicating the lack of any significant trend when no inversion occurs; and  
(b): easy-to-hard (E2H) and hard-to-easy (H2E) transitions show that, on average, there is a sharp and significant increase and decrease in loss patterns as samples are inverted to hard and easy difficulty groups respectively. Since SL does not directly take into account trend information, these results show that trend dynamics can inform our technical objective of developing better curricula. 
%In an extreme case, consider two training samples $i$ and $j$ with the exact same loss at current epoch but different loss trends, e.g., $i$ may have a rising trend and is about to become harder than before for the model while $j$ can have a falling trend and is about to become easier for the model. SL will assign the same weight to these samples, regardless of their loss trajectories. However, Trend-SL allows distinguishing such examples by taking into account loss trends.      

% The plots for E2H and H2E are near similar because based on hyper-parameter search, we had best validation performance for small value of alpha which is 0.1 for \gdpr, 0.7 for \pgr{} and for Wikipedia dataset. As the value of the alpha is smaller, the plots for both curricula follow closely. However, as shown in Table~ref{tab:performance}, capturing trend information is promising resulting in distinguishable performance.

% \cut{
% \begin{figure*}[!t]
%     \centering
%     \subfigure[Easy to Easy (E2E)]{\includegraphics[scale=0.25]{images/Easy to Easy trend_unit_variance_False_l2.png}\label{fig:e2e}}
%     \subfigure[Easy to Hard (E2H)]{\includegraphics[scale=0.25]{images/Easy to Hard trend_unit_variance_False_l2.png}\label{fig:e2h}}
%     \subfigure[Hard to Easy (H2E)]{\includegraphics[scale=0.25]{images/Hard to Easy trend_unit_variance_False_l2.png}\label{fig:h2e}}
%     \subfigure[Hard to Hard (H2H)]{\includegraphics[scale=0.25]{images/Hard to Hard trend_unit_variance_False_l2.png}\label{fig:h2h}}
%     \caption{}
%     \label{fig:transition}
% \end{figure*} 
% }

\begin{figure}[!t]
    \centering
    \subfigure[Easy to Hard]{\includegraphics[scale=0.18]{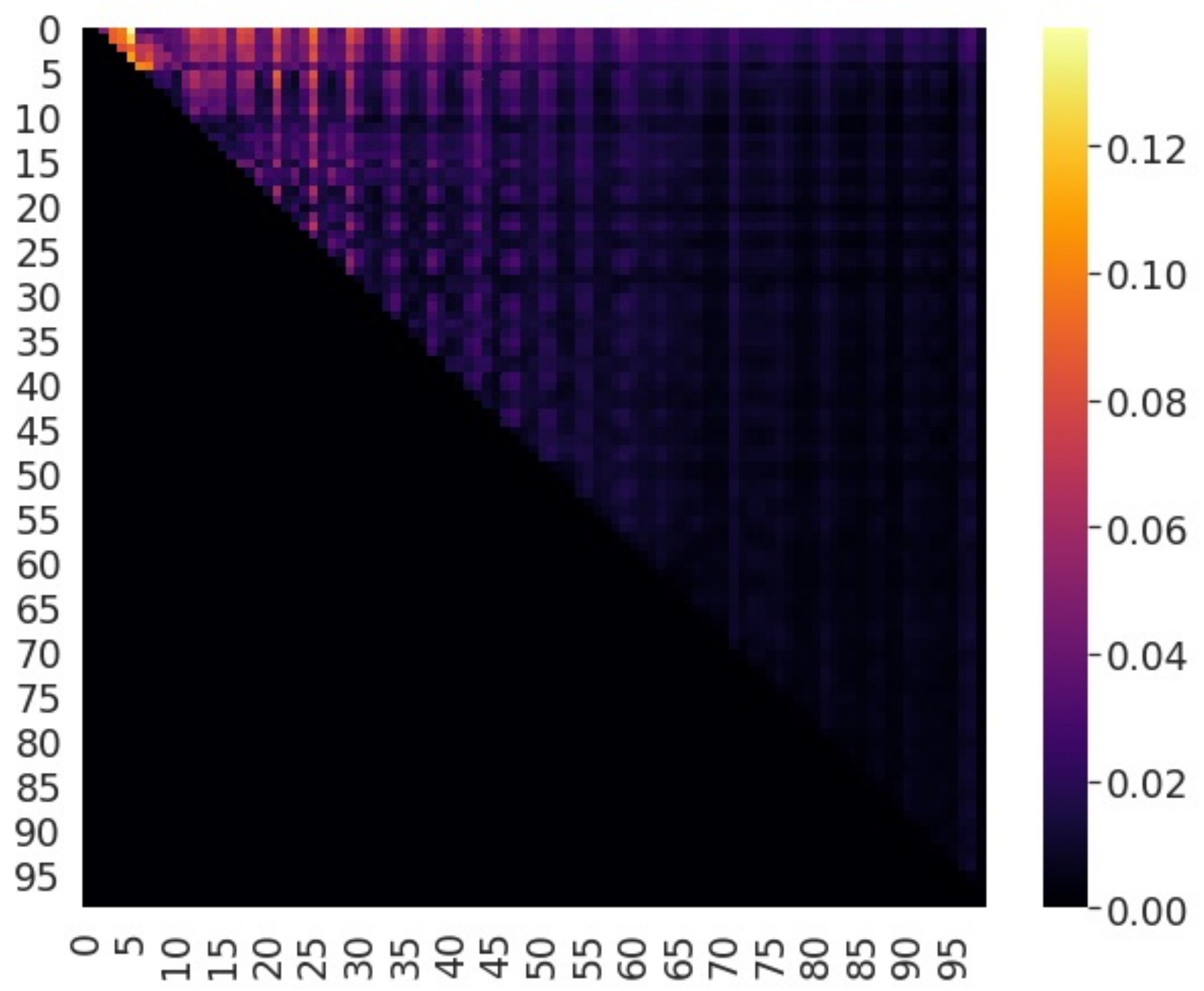}\label{fig:e2h_fraction}}
    \subfigure[Hard to Easy]{\includegraphics[scale=0.18]{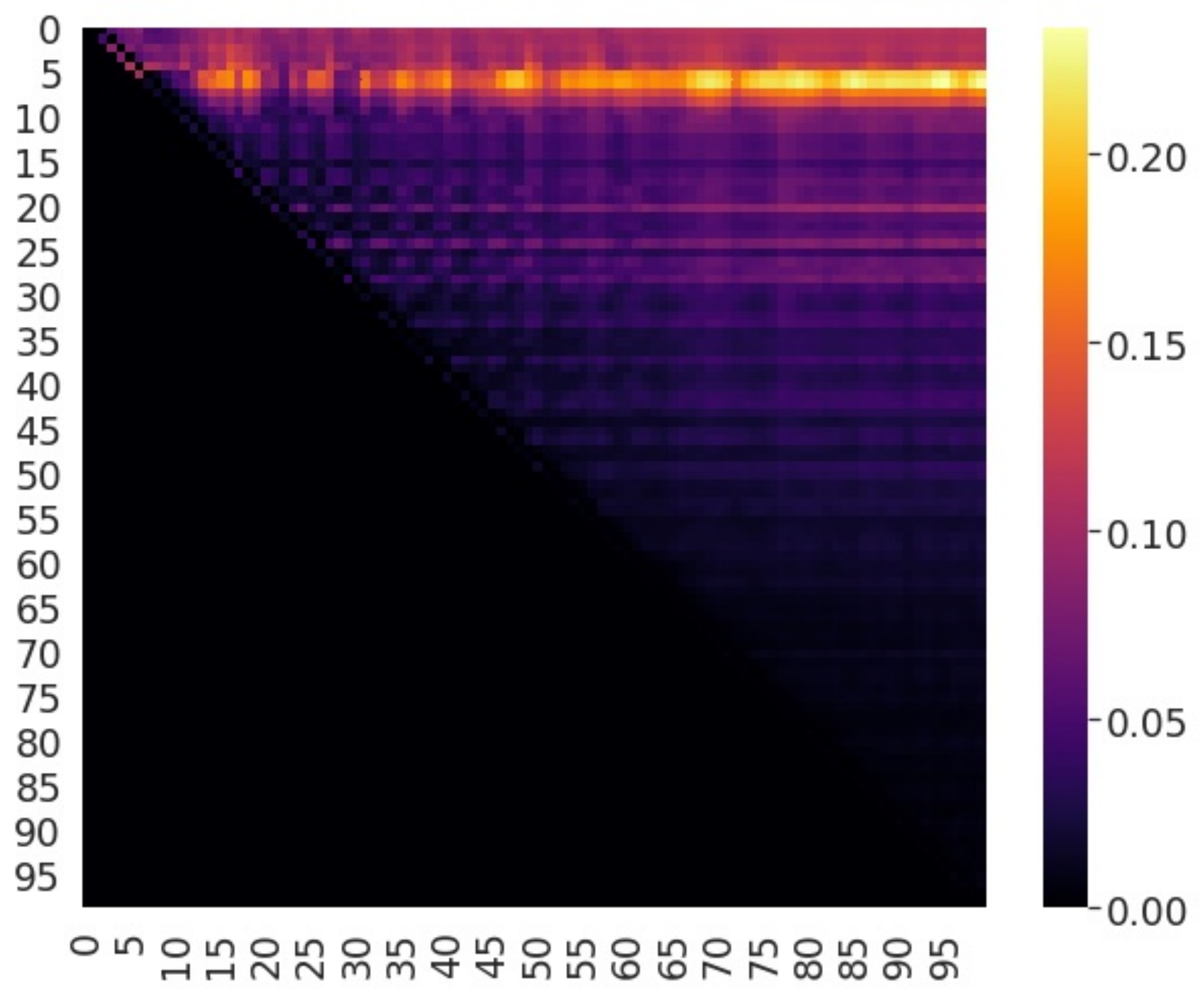}\label{fig:h2e_fraction}}
    \caption{Inversion dynamics at difficulty level during training: (a) inversions from easy to hard with rising loss trends and (b) inversions from hard to easy with falling loss trends. The initial epochs on the y-axis are brighter then later epochs, indicating that most inversions occur early in training.}
    \label{fig:fraction2d}
\end{figure}

\begin{figure*}[!t]
    \centering
    \subfigure[Easy to Hard]{\includegraphics[scale=0.4]{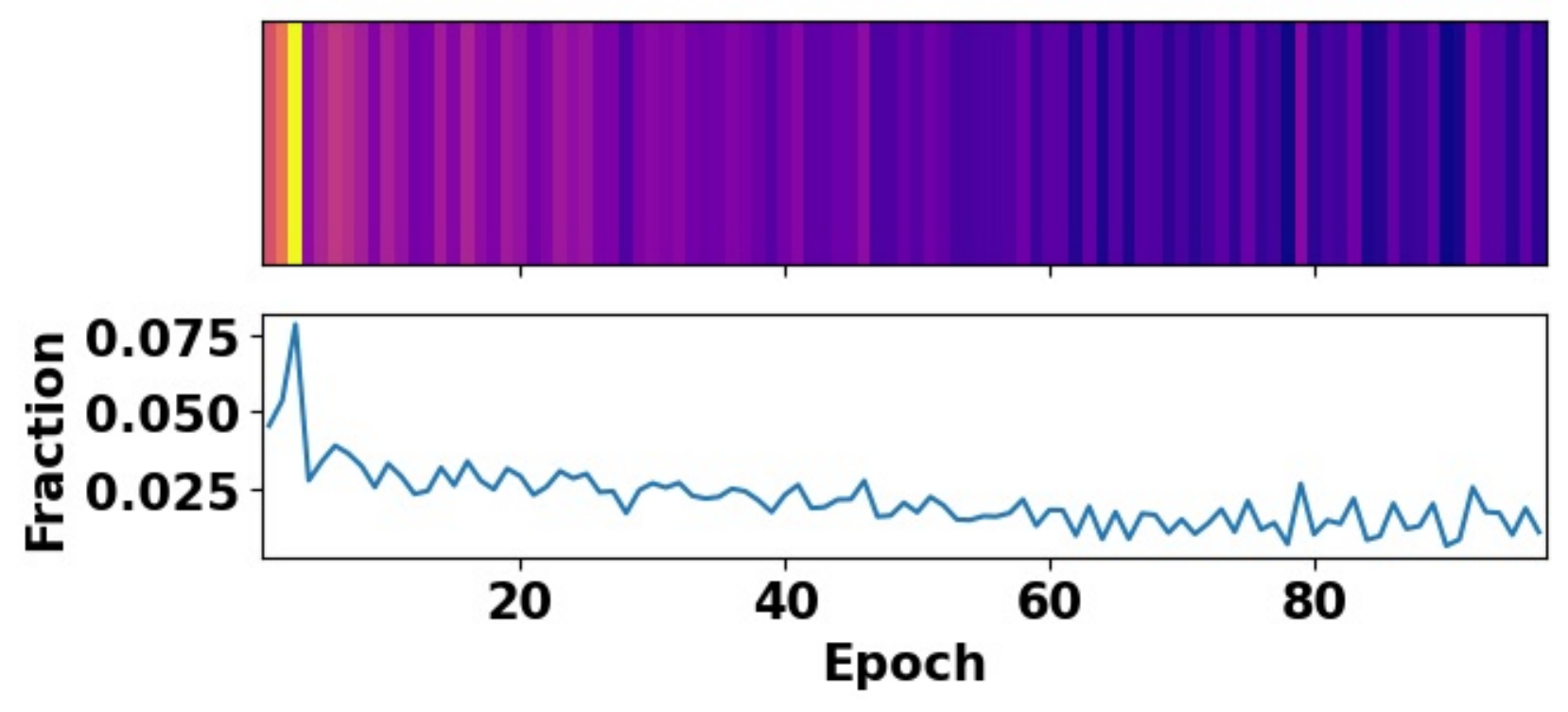}\label{fig:e2h_fraction_1d}}
    \subfigure[Hard to Easy]{\includegraphics[scale=0.4]{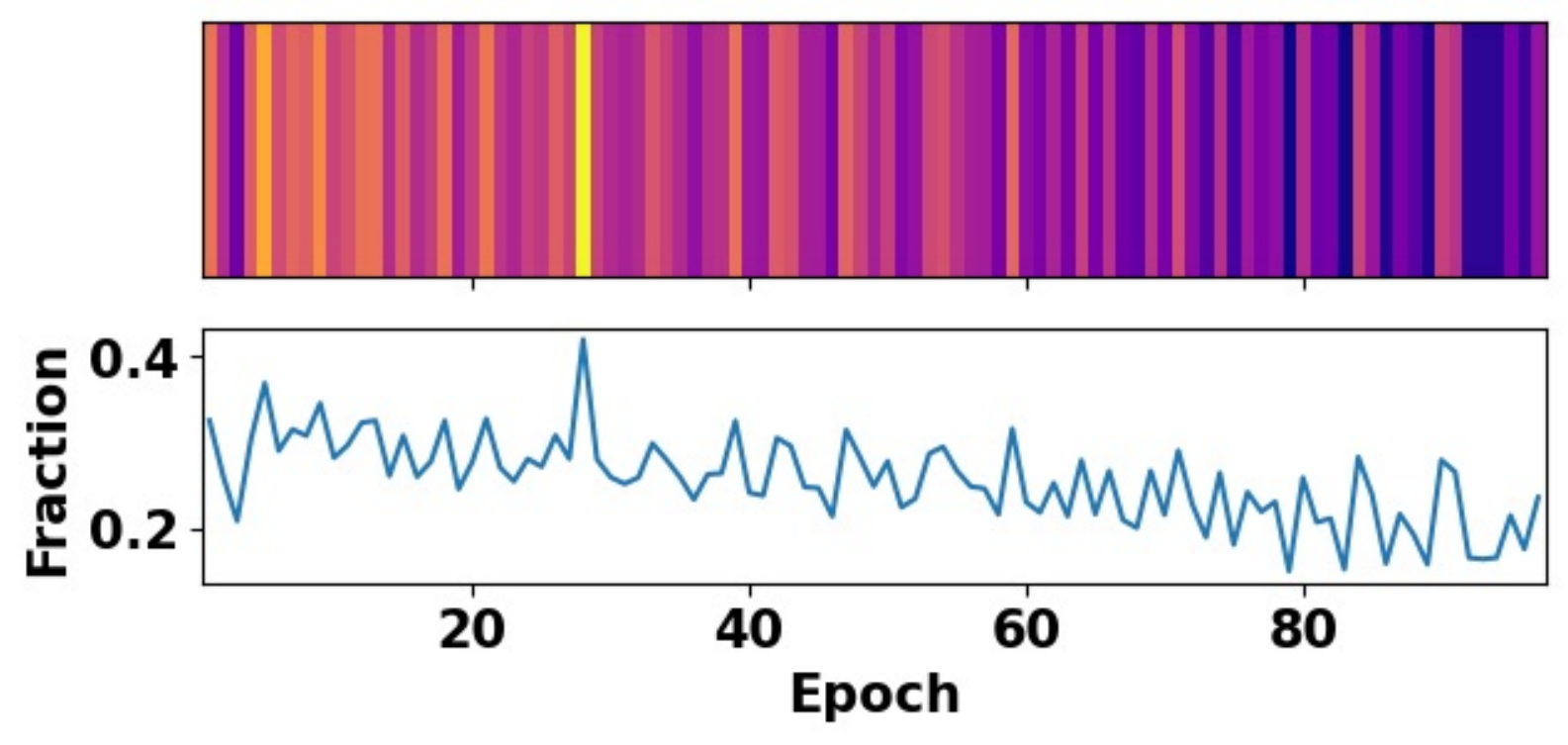}\label{fig:h2e_fraction_1d}}
    \caption{Inversion heatmap when (a): easy samples with rising loss trend become hard (left) and (b): hard samples with falling loss trend become easy (right).}
    \label{fig:fraction1d}
\end{figure*} 

\paragraph{Inversions occur early during training:}
Figure~\ref{fig:e2h_fraction} shows the fraction of samples that were easy at epoch $i$ but became hard with a rising trend at epoch $j$ > $i$. The corresponding heatmap for Hard-to-Easy with falling trend is shown in Figure~\ref{fig:h2e_fraction}. In both case, the initial epochs (see the y-axis) are brighter then later epochs, indicating that most inversions occur early in training and the effect of trend is more prominent in the initial part of training.
% This is expected as model behavior is initially affected by its initialization, stochastic gradient descent dynamics and random batch sampling. 
% We observe that most of the hard examples are converted to easy examples in the initial training epoch indicating the effect of trend is more prominent in the intial part of training and later as each of the example converges there is no major change in the loss happening and as result the change in the trend is minimal resulting is no change in difficulty cateogory and model becomes stable and does not learning anything new. Similarly for Easy-to-hard \ref{fig:e2h_fraction} heatmap.

\paragraph{Inversions occur with falling or rising loss trends:}
SL does not use trend information. However, its estimated difficulty for a considerable fraction of samples (with falling or rising loss trends) is inverted during training. In fact, we observe that 21.2\% to 50.0\% of hard samples that have a falling loss trend will become easy in their next training iteration; similarly 1.3\% to 11.1\% of easy samples that have a rising loss trend will become hard in their next training iteration. 
% auc for easy with rising trend converting to hard : 24.87
% auc for hard with falling trend. converting to easy : 4.51
% \paragraph{Falling or rising loss trends result in inversions:}
Figure~\ref{fig:fraction1d} shows the inversion heatmap for such Easy-to-Hard and Hard-to-Easy transitions in consecutive epochs.
% where the y-axis (Fraction) indicates the fraction of easy examples with a rising trend at epoch $e$ that became hard at epoch $e+1$. Similarly for Hard-to-Easy inversion heatmap. In Hard-to-Easy inversion the examples were hard at current epoch but had falling trend to eventually be inverted to easy examples. 
The area under the curve for Easy-to-Hard with rising trend and Hard-to-Easy with falling trend are 24.87 and 4.51 respectively. 
Trend-SL employs such trend dynamics to create better curricula.

% Fig. \ref{fig:fraction2d} show the number of instances that the examples changed the category from easy to hard and vice versa. For \ref{fig:e2h_fraction}, each row in the heatmap indicates the number of examples that were easy at epoch $i$ but where converted to hard and had rising trend at epoch $j$ > $i$. Similarly for Hard-to-Easy with falling trend. Higher the fraction brighter the heatmap. In Hard-to-Easy heatmap (Fig. \ref{fig:h2e_fraction}), region associated with initial epoch is relatively brighter then later epoch indicateing that most the examples which are considered hard are classified to easy in the initial epoch leaving more darker region for later epoch. Moreover, most of the hard examples are converted to easy examples in the initial training epoch indicating the effect of trend is more prominent in the intial part of training and later as each of the example converges there is no major change in the loss happening and as result the change in the trend is minimal resulting is no change in difficulty cateogory and model becomes stable and does not learning anything new. Similarly for Easy-to-hard \ref{fig:e2h_fraction} heatmap.

\subsection{Domain and Feature Analysis}
\begin{figure}[t]
    \centering
    \includegraphics[scale=0.3]{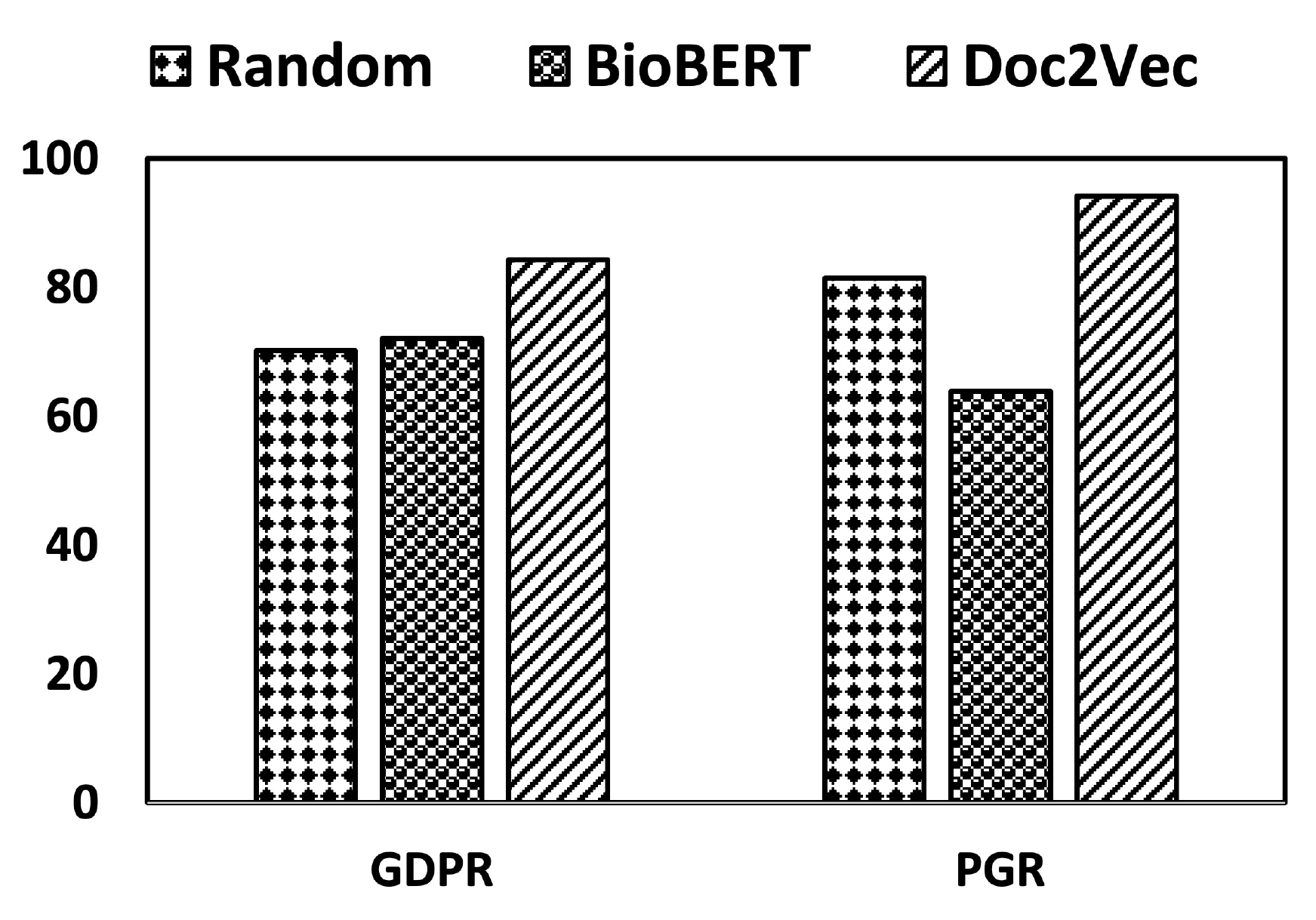}
    \caption{Performance of \gtnn{}  with Trend-SL with additional features.}
    \label{fig:ablation_w_feature}
\end{figure}

\paragraph{In-domain embeddings improve the performance:}
In these experiments, we re-train our model %without additional text features but 
with different embedding initialization. As shown in Figures~\ref{fig:ablation_w_feature}, Doc2Vec embeddings result in an overall better performance than BioBERT and random initialization approaches across the datasets. We attribute this result to in-domain training using text summaries of genes, diseases and phenotypes associated to {\em rare} diseases. In addition, the performance using BioBERT embeddings is either comparable or considerably lower than that of other embeddings including Random. This is perhaps due to pre-training of BioBERT using a large scale PubMED dataset, which has a significantly lower prevalence of publications on rare versus common diseases. On the other hand, we directly optimize Doc2Vec on in-domain rare-disease datasets, which leads to higher performance of the model. We tried to fine tune BioBERT on our corpus but as the text summaries are long, only a small fraction of texts (512 tokens) can be considered. 

\begin{figure}[t]
    \centering
    \includegraphics[scale=0.3]{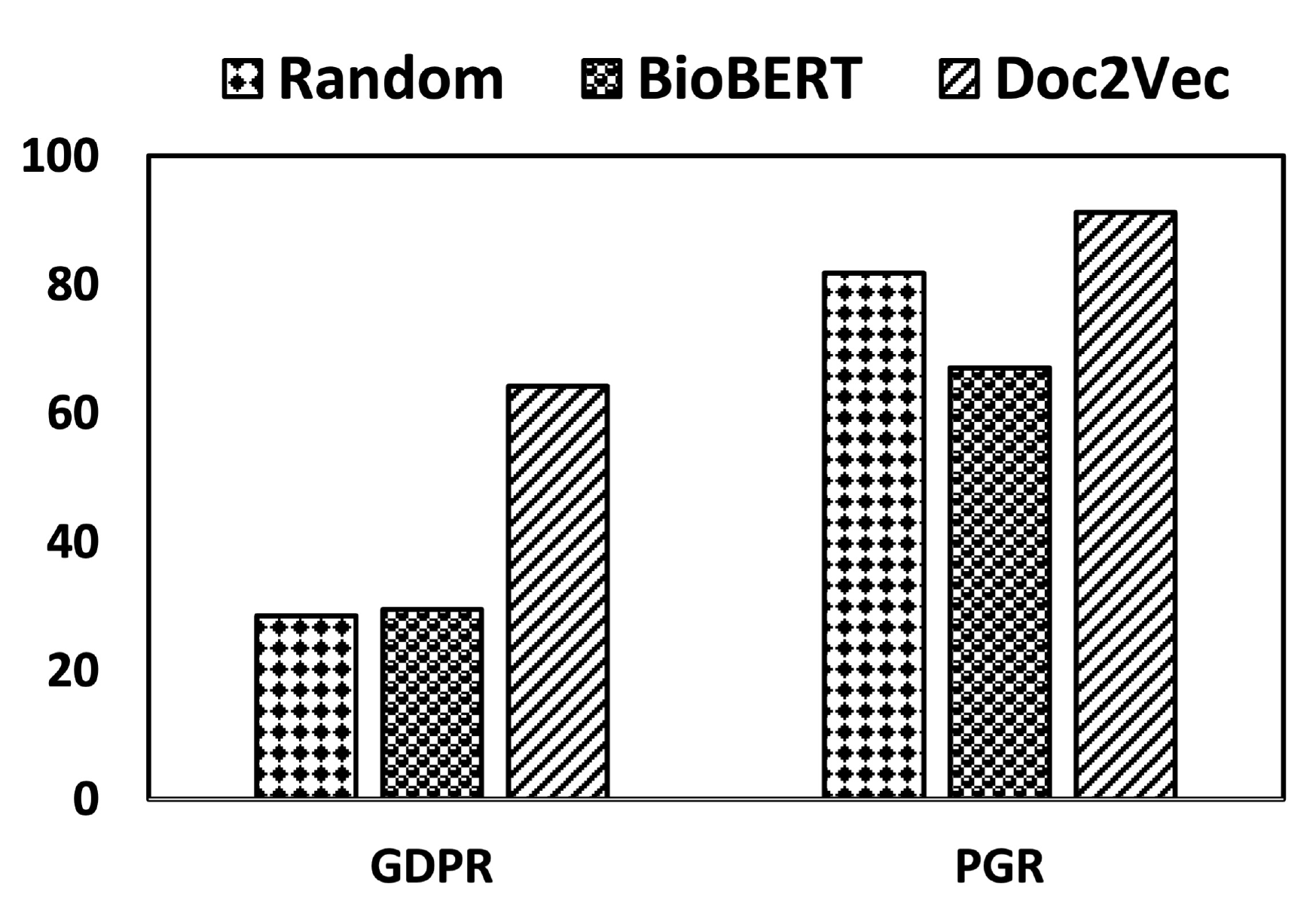}
    \caption{Performance of \gtnn{}  with Trend-SL without additional features.}
    \label{fig:ablation_wo_feature}
\end{figure}

\paragraph{Additional Features improve the performance:}
We re-train our models and exclude additional feature (i.e., relevance scores for \gdpr~and sentence embeddings for \pgr), with different node embedding initialization. 
Figure~\ref{fig:ablation_wo_feature} shows that excluding these features considerably reduces the F1-scores of our model across datasets and embedding initialization. 
These results show that both text features and information obtained from graph structure contribute to predicting relations between nodes. 

\section{Related Work} \label{relatedwork}
Previous research on relation extraction can be categorized into text- and graph-based approaches. In addition, to our knowledge, there is limited work on curriculum learning with graph datasets. 
% We describe these research work below.

%Previous research on relation extraction can be categorized into text-based or graph-based approaches with limited work on relation extraction on textual content using graph information. Moreover, to our knowledge, there are very few works on curriculum learning with graphs. We describe these research work below.

\paragraph{Text-based models:} Text-based methods extract entities and the relations between them from given texts.
Although, previous works typically focus on extracting intra-sentence relations for entity pairs in supervised and distant supervised settings~\cite{sousa2019silver,mintz-etal-2009-distant,dai2019distantly,lin-etal-2016-neural,peng2017cross, zhang-etal-2018-graph,alex2017protein,zhang2018link,quirk-poon-2017-distant}, there are relation extraction approaches that focus on inter-sentence relations~\cite{kilicoglu2016inferring,yao-etal-2019-docred}. \citet{kilicoglu2016inferring} investigated multi-sentence relation extraction between chemical-disease entity pairs mentioned at multi-sentence level. They considered lexical features, and features obtained from intervening sentences as input to a classifier.
A close related work to our study has been conducted by~\citet{sousa2019silver}, who developed an effective model to detect relations between genes and phenotypes at sentence-level using sentential context and medical named entities in text. We compared our approach with \citet{sousa2019silver} on the dataset that they developed (PGR), see Section~\ref{sec:baselines}.

%Previous work extracted relations from intra and inter-sentence ~\cite{wang2019overview,sousa2019silver,thillaisundaram-togia-2019-biomedical,sahu-etal-2016-relation,dai2019distantly,kilicoglu2016inferring,yao-etal-2019-docred}. \citet{kilicoglu2016inferring} investigated multi-sentence relation extraction between chemical-disease entity pairs mentioned at multi-sentence level. They considered lexical features, and features obtained from intervening sentences as input to a classifier.

\paragraph{Graph based models:}
Previous research show that adding informative additional features with graph helps models learn better node representations for extracting relation between entity pairs. For example, \citet{zhang2018link} used distance metric information, and \citet{lidistance} used distance features like shortest path and landing probabilities between pair of nodes in subgraphs as additional features. We note that some graph properties, although informative and effective, can be expensive to calculate on large graphs during training and should be computed offline. 

\paragraph{Curriculum learning with graph data:}
Curriculum learning approaches design curricula for model training and generalizability~\cite{bengio2009curriculum,kumar2010self,Jiang2015-ek,amiri-etal-2017-repeat,jiang2018mentornet,castells2020superloss,zhou2020curriculum}. The common approach is to detect and use easy examples to train the model and gradually add harder examples as training progresses.
Curricula can be static and pre-built by humans or can be automatically and dynamically learned by the model. There are very few curriculum learning methods designed to work on the graph structure. \citet{wang2021curgraph} developed CurGraph, which is a curriculum learning method for sub-graph classification. The model estimates the difficulty of samples using intra and inter-class distributions of sub-graph embeddings and orders training instances to initially expose easy sub-graphs to the underlying graph neural network followed by harder ones. 
As opposed to static curriculum, \citet{saxena2019data} introduced a dynamic curriculum approach which automatically assigns a confidence score to samples based on their estimated difficulty. However, the model requires a large number of extra trainable parameters especially when data set is large. To overcome this limitation, \citet{castells2020superloss} introduced a framework with similar idea but calculates the optimal confidence score for each instances using a closed-form solution, thereby avoiding learning extra parameters. We extended this approach to include trend information at sample-level for learning effective curriculum.
% Similarly, \cite{xu-etal-2020-curriculum} pre-determines the difficulty of the text and use it to better train the model for classification.
%
%As opposed to static curriculum, \citet{saxena2019data} introduced an approach to automatically assign confidence score to dynamic curriculum to automatically assign a confidence score to each instance based on its estimated difficulty but requires to learn extra trainable parameters per instance. To overcome this limitation, \citet{castells2020superloss} introduced a framework with similar idea but uses closed-form solution to calculate optimal confidence score. 
\paragraph{Graph neural networks for NLP:}
There are several distantly related work that develop graph neural network algorithm for downstream tasks such as semantic role labeling~\cite{marcheggiani2017encoding}, machine translation~\cite{bastings2017graph,marcheggiani2018exploiting}, multimedia event extraction~\cite{liu2020story}, text classification~\cite{yao2019graph,zhang2020every} and abstract meaning representation~\cite{song2018graph}. Graph neural networks are used to model word-word or word-document relations, or applied to dependency trees. 
\citet{yao2019graph} generated a single text graph using word occurrences and document word relations from text data, and used the GCN method to learn embeddings of words and documents. Similarly, \citet{peng2018large}  used GCN to capture the semantics between non-consecutive and long-distance entities. 

% Compared to previous works, our approach uses multimodal information including both text and graph data as input. In addition, we use textual summaries as auxiliary inputs to the graph rather than modeling sentences as graphs.  Moreover, we are the first to extend \cite{castells2020superloss} framework and introduce novel model that uses curriculum to make better edge prediction using past information.

\section{Conclusion and Future Work}
We propose a novel graph neural network approach that effectively integrates textual and structural information and uses loss trajectories of samples during training to learn effective curricula for predicting relations between given entity pairs. Our approach can be used for both sentence- and document-level relation extraction, and shows a sizable improvement over the state-of-the-art models across several datasets. 
In future, we will investigate curriculum learning approaches for other sub-tasks of relation extraction, develop more effective techniques to better fit trends to time series data, and investigate the effect of curricula on other graph neural networks for relation extraction. 
% Due to the length of the document, our model currently uses embeddings obtained from in-domain training. In future, we plan to develop an approach to effectively understand the semantics of the long document using transformer based language model.
% We further improve the model by incorporating the curriculum learning.

\bibliography{custom}
\bibliographystyle{acl_natbib}
\section*{Ethical statement}
This investigation partially uses data from the field of medicine. Specifically, it includes genes, diseases and phenotypes that contribute to rare diseases. Although the present work does not include any patient information, it is translational in nature and its broader impacts are first and foremost the potential to improve the well-being of individual patients in the society, and support clinicians in their diagnostic efforts, especially for rare diseases. Our work can also help Wikipedia curators and content generators in finding relevant concepts.

%\section{Appendix}
%\label{sec:appendix}

%\input{sections/supplementry.tex}

\end{document}